\documentclass[letterpaper, 10 pt, conference]{ieeeconf}  

\IEEEoverridecommandlockouts                              

\usepackage{graphicx} 
\usepackage{amsmath} 
\usepackage[linesnumbered, ruled]{algorithm2e}
\usepackage{enumerate}
\usepackage{gensymb}
\usepackage[font={footnotesize}]{caption}
\usepackage[font={footnotesize}]{subcaption}
\usepackage{comment}
\usepackage{CJKutf8}
\usepackage{booktabs}

\usepackage{color}
\usepackage[dvipsnames]{xcolor}
\definecolor{myRed}{HTML}{CF4858}
\definecolor{myGray}{HTML}{4D505B}
\definecolor{myGreen}{HTML}{16A79D}
\definecolor{myPurple}{HTML}{80628B}
\definecolor{myOrange}{HTML}{F4AC42}

\ifodd 0
\newcommand{\chiyu}[3][Brown]{{\color{blue}\{C.D-} {\color{#1}{#2}} {\color{blue}[COMMENT: #3]-C.D\}}}

\newcommand{\chen}[3][myOrange]{{\color{orange}\{C.F-} {\color{#1}{#2}} {\color{orange}[COMMENT: #3]-C.F\}}}

\else
\newcommand{\chiyu}[3][black]{{\color{black}} {\color{#1}{#2}} }
\newcommand{\chen}[3][black]{{\color{black}} {\color{#1}{#2}} {}}

\fi

\title{\LARGE \bf
Low-cost LIDAR based Vehicle Pose Estimation and Tracking
}

\author{Chen Fu$^1$, Chiyu Dong$^1$, Xiao Zhang $^1$ and John M. Dolan$^{1,2}$ 
\thanks{$^1$Chen Fu, Chiyu Dong and Xiao Zhang are with the Department of Electrical and Computer Engineering, Carnegie Mellon University, 
          Pittsburgh, PA 15213, USA
        {\tt\small \{cfu1, chiyud, xiaozhan\}@andrew.cmu.edu}}%
\thanks{$^2$John M. Dolan is with the Department of Electraical and Computer Engineering and the Robotics Institue, Carnegie Mellon University, 
         Pittsburgh, PA 15213, USA
        {\tt\small jmd@cs.cmu.edu}}%
}

\begin{document}

\maketitle
\thispagestyle{empty}
\pagestyle{empty}

\begin{abstract}
\chiyu{
Detecting surrounding vehicles by low-cost LIDAR has been drawing enormous attention. In low-cost LIDAR, vehicles present a multi-layer L-Shape. Based on our previous optimization/criteria-based L-Shape fitting algorithm, we here propose a data-driven and model-based method for robust vehicle segmentation and tracking. The new method uses T-linkage RANSAC to take a limited amount of noisy data and performs a robust segmentation for a moving car against noise. Compared with our previous method, T-Linkage RANSAC is more tolerant of observation uncertainties, i.e., the number of sides of the target being observed, and gets rid of the L-Shape assumption. In addition, a vehicle tracking system with Multi-Model Association (MMA) is  built upon the segmentation result, which provides smooth trajectories of tracked objects. A manually labelled dataset from low-cost multi-layer LIDARs for validation will also be released with the paper. Experiments on the dataset show that the new approach outperforms previous ones based on multiple criteria. The new algorithm can also run in real-time. -}{}
\chen{}{Can we mention tracking as a new pipeline? single model vs selection from multiple model}

\end{abstract}

\section{INTRODUCTION}
\chiyu{
Autonomous driving cars have been drawing tremendous interest in the academy, industry and marketing. Self-driving cars for end users are edging closer, as advances in the technology improve their ability to drive efficiently and safely in traffic. For example, Uber and Waymo released their Beta version self-driving car-sharing services in limited areas across the U.S. Leading car makers such as GM, Ford and Tesla have shipped  Advanced Driving Assistance Systems with their production cars, which enable the cars with active safety functions or ``Level-3" autonomy. %

To provide the car with a certain level of autonomy and safety, a perception system that robustly understands the surrounding environment is crucial. It provides important references for decision making and planner modules which directly decide the behaviors of a self-driving car in different scenarios. However, the high cost of the perception system limits the commercialization potential of autonomous driving cars. %
There is a trade-off in the perception system: even though high-end sensors (e.g., Velodyne's HDL-64) easily provide a clear understanding of the environment, their high expense is prohibitive for consumers. On the other hand, low-cost sensors require more sophisticated algorithms to achieve an acceptable level of sensing. However, there is lack of development on the method of segmentation and tracking for a multi-layer low-cost LIDAR in autonomous driving scenarios. %
The algorithms for the multi-layer low-cost LIDAR should be able to segment and track vehicles with much less and more noisy LIDAR points than that of high-performance LIDAR. Besides the sparsity and uncertainty of points, incomplete observations, such as partial L-shapes should be also handled.
}{}

\chiyu{
In this paper, we introduce a novel approach to perform vehicle segmentation and tracking using low-cost, commercialized multi-layer 2D LIDAR. To segment a vehicle from a sparse point cloud,  T-linkage RANSAC is applied. T-linkage RANSAC inherits the advantages of the classic RANSAC algorithm. In addition, it does not require prior knowledge of the number of models, e.g., number of the best fit lines in a cluster of points. Especially in our application, L-shapes sometimes are partially observed due to relative positions, so there is no guarantee that the ego-vehicle always observes a specific number of sides of the target vehicle. Without a manual setup or assumption, T-Linkage RANSAC automatically determines whether it needs to fit one line or an L-shape to a cluster of points.  The goals of the task are clear: 1) Fitting vehicles with rectangular bounding boxes, 2) Identifying different vehicles with consistent tracker IDs, 3) Generating smooth historical trajectories for each tracked target. This approach plays a significant role on our self-driving platform, as its results are used and affect performance at different levels of planning. The advantages of the new method are tied to the features of low-cost LIDAR:  It processes spatially sparse observations with fewer assumptions on the shape of the point cluster and the results are robust against noise in the distance measurements. 
}{The description of the advantages can be more technically specific.}

\chiyu{
The organization of this paper is as follows: Section \ref{sec:related} briefly reviews related work and the state of the art in 2D LIDAR segmentation and tracking. Section \ref{sec:pose} introduces the segmentation procedure. Section \ref{sec:fitting} describes a method for finding the best bounding box for a segmentation. Sections \ref{sec:association} and \ref{sec:tracking} discuss tracking. Finally, sections \ref{sec:result} and \ref{sec:conclusion} give experimental results and conclude with future work. }{}






\section{Related Work}\label{sec:related}
\chen{
In 2007, the autonomous vehicle ``Boss'' developed by Carnegie Mellon University (CMU) won the DARPA Urban Challenge, which is a historical success for autonomous driving. To accurately detect and track obstacles and vehicles in the competition,  a high-end Velodyne HDL-64 sensor was installed on ``Boss'' \cite{darms2009obstacle}. However, this type of non-automotive-grade sensor is not affordable for the average consumer. The Cadillac SRX platform is another milestone of CMU autonomous vehicle research \cite{wei2013towards}. With only minimal modification of its appearance, multiple automotive-grade low-cost sensors including six 4-layer LIDAR, RADAR and cameras have been integrated into the vehicle \cite{cmuplatform}. To guarantee a similar level of driving performance without high-end sensors, a more robust and state-of-the-art perception system is required.  
}{}

\chen{The most popular vehicle tracking approach can be summarized into three steps: data segmentation, data association and Bayesian filter-based vehicle state estimation \cite{Petrovskaya2009}. During the first stage, the data points are separated into meaningful clusters and different vehicle models such as perpendicular corner, L-shape and rectangle box\cite{maclachlan2006tracking, shen2015efficient,mertz2013moving}. To estimate the pose and heading angle of the vehicle, many model-based fitting methods have been developed. In \cite{shen2015efficient}, the data points are divided into two sets, which are fitted to two perpendicular lines. However, this method requires the sequencing information of the scanning points, which limits the scalability of the algorithm. Instead of splitting data points into two sets, Principal Component Analysis (PCA) can be applied directly to the whole cluster to select the principal axes. The pose of the vehicle can be estimated using these axes\cite{zhao2009moving}. In \cite{zhang2017efficient}, Zhang proposed a search-based method to find the heading angle of the vehicle. Multiple criteria are applied to evaluate the optimization-based fitting. 
In our paper, a T-linkage RANSAC-based best-fitting selection method is proposed for the first stage. Taking advantage of the T-linkage algorithm, outliers including failures of low-cost sensors and cross-layer overlapping can be eliminated. In addition, no assumption is made about the number of model instances.}{}

\chen{In the second step, accurate data association is essential for the tracking process. In both LIDAR and vision-based tracking tasks, Multiple Hypothesis Tracking (MHT) is a promising solution. In MHT, the possible track hypotheses for every candidate track form a tree structure. To select the best track hypothesis and prune other invalid hypotheses, a likelihood of each track is calculated accordingly \cite{mht2015}. In contrast, the joint probabilistic data association (JPDA) method considers all the potential hypotheses. In each time frame, a joint probability score is calculated to associate the measurement with confirmed tracks \cite{jpda2015}. MHT and JPDA are widely used, but they suffer from the similar limitation that the computation complexity is relatively large for on-board computation. In our proposed method, we view the data association step as an integer programming problem. In order to solve the problem of the deformation and missing objects caused by occlusions and sparsity of the low-cost sensor, we keep a list of previously confirmed trackers and apply the Hungarian algorithm to find the optimal assignment.}{}

\chen{In the final step, the vehicle state estimation, a Kalman Filter (KF), extended Kalman filter (EKF) or Unscented Kalman Filter (UKF) is commonly applied. In \cite{MMM2017}, a single vehicle state model with KF has been applied to vehicle tracking using low-cost LIDAR.  In \cite{1709.08517}, the Independent Steered Model (ISM) and Variable-Axis Ackerman Steered Model (VASM) have been combined with KF to predict the state of the moving vehicle in each frame. In \cite{Ego2008}, moving vehicles can be detected as the outliers of ICP methods, with EKF applied to estimate the non-linear state of each vehicle. All these methods are based on the assumption that the vehicle model and the co-variance of the environment, which might be inaccessible, are well known. Instead of using the three-step tracking framework, other types of techniques are also proposed. In \cite{Petrovskaya2009}, a particle filter-based method is used which directly tracks the vehicles without the need of the data segmentation and association steps. However, how to improve the speed and efficiency of this family of methods is still an open problem. 
In this paper, we apply the multiple model association (MMA) algorithm, which considers the co-variance of pre-fit residual error to select the best vehicle model and KF parameter setting. It eliminates the effect of the inaccuracy caused by the assumption of single-vehicle models, without dramatically increasing the computation complexity.}{}

\section{Vehicle Pose estimation and tracking}\label{sec:pose}

\subsection{RANSAC-based L-Shape Fitting Pipeline}\label{sec:fitting}

\begin{figure}
      \centering
    \begin{subfigure}[t]{0.45\linewidth}
      \centering
      \includegraphics[width=1\linewidth]{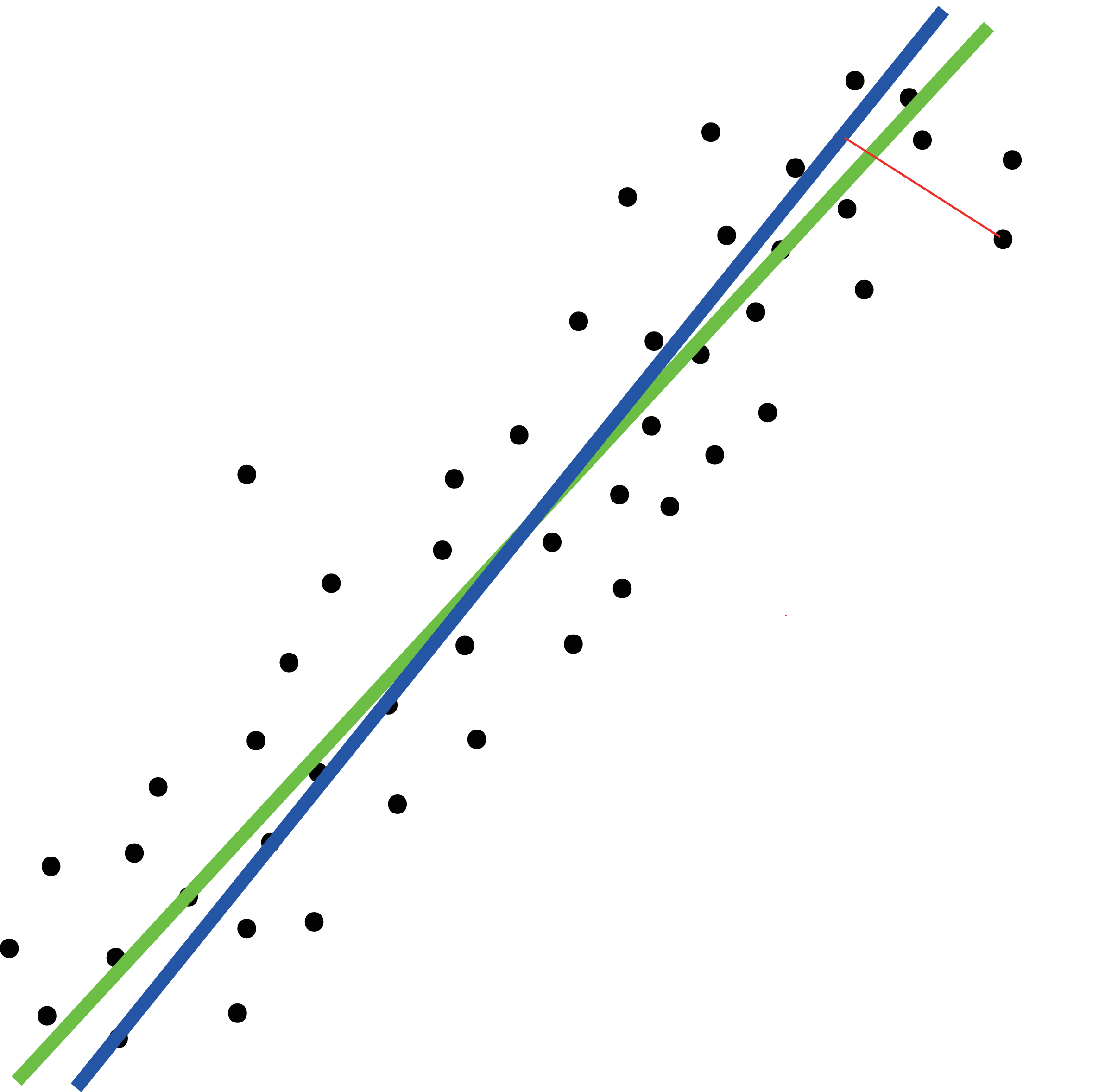}
      \caption{Predicted line and target line with smaller variance of absolute residual error} \label{fig:fitting_ill1}
    \end{subfigure}
    \hfill
    \begin{subfigure}[t]{0.45\linewidth}
        \centering
        \includegraphics[width=1\linewidth]{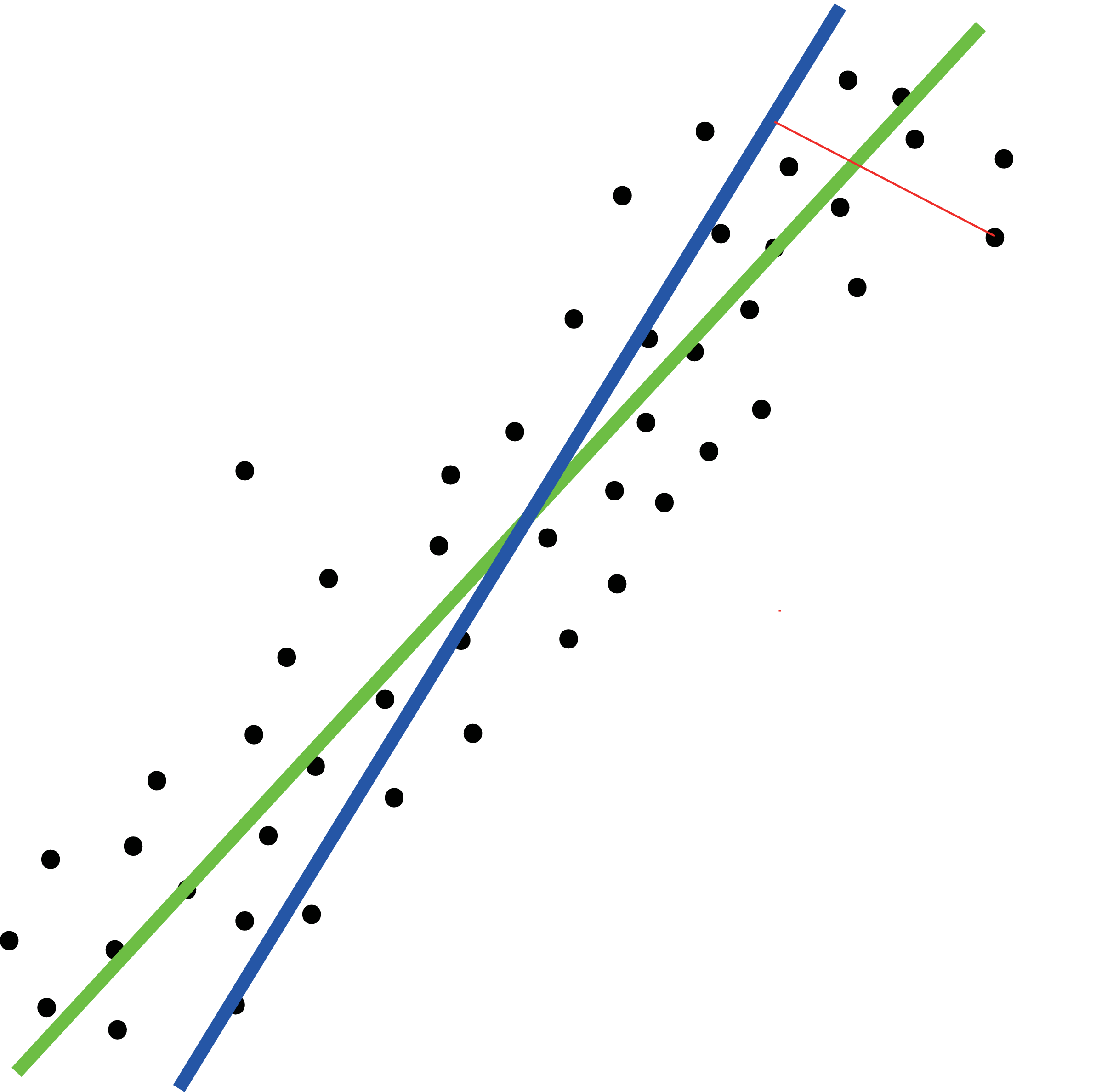} 
        \caption{Predicted line and target line with larger variance of absolute residual error} \label{fig:fitting_ill2}
    \end{subfigure}
    \caption{This figure illustrates the relationship between the variance of the absolute residual error and correctness of the fitted line. The green line indicates the target line, the blue line indicates the predicted line and the red line illustrates the residual error or distance between data points and predicted line. In (a), the predicted line better fits the data points, and the data points are more evenly distributed along the predicated line. Therefore, the variance of the absolute residual error of the predicted line in (a) is smaller than (b).}
    \label{fig:fitting_ill}
   \end{figure}

An accurate understanding of the surrounding environment is critical for autonomous vehicles. However, previous L-shape methods have drawbacks including the requirement of sequence information for the scanning points as well as sensitivity to outliers and error data points. Noise might be caused by the uncertainty of the environment, failure of the sensor, or deformation of the vehicle observation. In order to overcome these limitations, we designed a new L-shape fitting pipeline by applying a random sampling mechanism to get rid of outliers. To further improve the accuracy, a variance of the residual error cost function is developed to choose the best-fit vehicle pose. The residual error is defined as the distance between data points and predicted lines (see Fig. \ref{fig:fitting_ill}). 

\begin{figure}[tbp]
      \centering
      \includegraphics[width = \linewidth]{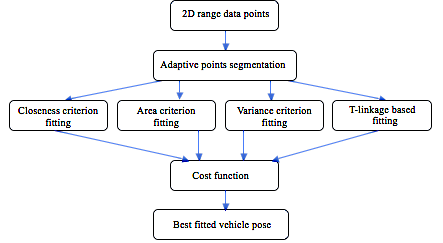}
      \caption{The pipeline of the proposed L-shape fitting algorithm.}
      \label{fig:1}
\end{figure}

The proposed method is four-fold: adaptive scanning points segmentation, T-linkage-based outlier removal and L-shape points clustering, rectangle fitting using different criteria, and cost function-based best fit selection. The pipeline is summarized in Fig. \ref{fig:1}.

   The conventional RANSAC method is robust when dealing with a single-model instance corrupted by outliers. However, it cannnot tolerate either true outliers caused by random noise or pseudo-outliers caused by other model instances \cite{jlinkage}. The multi-RANSAC algorithm needs prior knowledge of the number of instances, which is not generally available \cite{Zuliani05themultiransac}. As a consequence, we adopt the T-linkage method, which does not require assumptions about the number of instances and is more robust to all kinds of outliers. 
   
In order to cluster range data points, a set of model hypotheses is generated randomly, which in our case is a set of lines. We define the preference function (PF) of model hypotheses $ H = \{h_i\}_{i = 1 ... m}$ with the inlier threshold $\tau$ as \cite{Magri14t-linkage:a}:

$$\phi_{i} = 
    \begin{cases}
      e^{-d(h_{i}, x_{j}) / \tau}, & \text{if}\ d( h_{i}, x_{j}) < \tau, \\
      0, & \text{if}\ d(h_{i}, x_{j}) \geq \tau    
     \end{cases}
     \eqno{(1)}
$$

The contribution of the T-linkage algorithm compared to other RANSAC-based methods is that, instead of using a $\{0, 1\}$ indicator of overlapping as in the Jaccard distance, the distance is extended to the Tanimoto distance in continuous space $[0, 1]$, which is a relaxation of binary assignment \cite{Tdist}:

$$
d_{T}(p, q) = 1 - \frac{<p, q>}{||p||^2 + ||q||^2 - < p, q>} \eqno{(2)}
$$

Using the T-linkage algorithm, we can overcome not only the true outliers caused by random noise, but also pseudo-noise caused by cross-layer overlapping and vehicle deformation. To determine the heading of the vehicle, we define the cluster with the largest number of range data points as the dominant cluster. The slope of the line w.r.t. the dominant cluster reveals the heading orientation of the vehicle. The proposed T-linkage-based outlier removal and vehicle pose estimation algorithm is presented in Alg. \ref{alg:1}. 
\begin{algorithm}[t]
\SetKwInOut{Input}{Input}
 \SetKwInOut{Output}{Output}
 \Input{Segmented 2D range data points}
 \Output{Cluster of points belonging to same label and vehicle heading orientation}
 Label each data point as its own model hypothesis and calculate corresponding PF $\phi_{i}$ using (1)\;
 \While{$\exists$ $i$ such that $\phi_i$ not orthogonal}{
     Find $p, q$ such that $d_T (p, q) = \min_{m,n} d_T(m, n)$\;
     Merge cluster $p, q$ and compute the PF of the resulting new cluster
     }
     Outlier detection and removal\;
     Find the dominant cluster\;
     Heading angle estimation by line fitting\; 
 \caption{T-linkage-based outlier removal and vehicle pose estimation}
 \label{alg:1}
\end{algorithm}

The conventional single-criteria fitting algorithm estimates L-shape using all the input range data points and projects all the points onto a particular direction. The criterion also considers the distribution of points around the two on-line projection boundaries. As a result, it loses another dimension of information, which is the residual error from the data points to the proposed fitting line. This explains why a single scan point of the side mirror can affect the fitting result. Taking advantage of the previous T-linkage clustering, we can directly estimate the quality of each fitting rectangle using the dominant cluster. Therefore, we designed a cost function to evaluate the variance of the residual error, which calculates the score of each rectangle from different fitting criteria and the T-linkage-based method. A lower variance of the absolute residual error indicates that the predicted line is more parallel to the target line, as shown in 
$$
   L(H) = Var(|H \times x_{i}|) ~~for~all~x_{i} \in \{X\}
   \eqno{(3)}
$$
where $H = [a ~b ~c]$ defines the estimated vehicle orientation and $X$ is the points in the dominant cluster.

\subsection{Data Association using Hungarian Algorithm}\label{sec:association}
In the vehicle tracking process, the data association is usually the most difficult problem due to occlusion, deformation and missing targets. To face these challenges, instead of using the global nearest neighbor (GNN) approach, we define this scenario as a multi-target-multi-observation data association problem which can be formulated as a binary integer program with binary matrix $A$ and score matrix $S$ with the dimension of $n \times n$:
$$
\begin{aligned}
& \text{Minimize}
& & \sum_{h = 1}^{n}\sum_{k = 1}^{n}A_{hk}S_{hk} \\
& \text{Subject to}
& & \sum_{h = 1}^{n} A_{hk} = 1, \; k = 1, \ldots, n. \\
&&& \sum_{k = 1}^{n} A_{hk} = 1, \; h = 1, \ldots, n. \\
&&& A_{hk} \in \{0, 1\}
\end{aligned}
\eqno{(4)}
$$
The above problem has a worst-case complexity which is NP-hard. We can instead solve it as an assignment problem with the Hungarian algorithm in polynomial time. The assignment algorithm will assign all the observations, which might include false detection with corresponding trackers. As a result, to prevent false assignment, a gate function should be applied. According to the Kalman filter (KF), the potential observation at a particular frame should be located in a gate area determined by the mean $\hat{x}$ and co-variance $\hat{P}$ of the tracker \cite{Reid79analgorithm}. Here we define $H$ as the system observation matrix, $z_{t}$ as the observation at time frame $t$, and $R$ to be the co-variance of observation noise.
$$
\begin{aligned}
    (z_{t} - H\hat{x})B^{-1}(z_t - H\hat{x}) < \epsilon, \\
    B = H\hat{P}H^{T} + R
\end{aligned}
\eqno{(5)}
$$

\subsection{Multiple Model Vehicle Tracking}\label{sec:tracking}
The low-cost LIDAR has the limitation that the data points may not be stable across each scan and it is difficult to estimate the size directly from the observation. As a result, instead of tracking the center of the vehicle, which requires consideration of the geometry of the vehicle, we track the nearest corner of the vehicle, which is more stable. The motion of the surrounding vehicles is essential information for the ego-vehicle, which can be represented by vehicle state vectors. The state vector $X$ contains the vehicle position, velocity, acceleration, yaw angle and yaw rate. 
    $$
      X = (x, y, v_x, v_y, a_x, a_y, \theta, \delta)^T
      \eqno{(6)}
    $$
The dynamics of a vehicle can be summarized as different vehicle models. Most on-road vehicles can be modeled by the following three types of dynamic vehicle model:
\begin{enumerate}
    \item \textit{Stationary Vehicle Model}:
    In this model, velocity, acceleration, yaw angle and yaw angle velocity are set to zero. In this case, the current location of the vehicle only depends on the previous state, so the state vector can be simplified to $X= (x, y, \theta)^T$. The roadside parking vehicles and vehicles waiting at the intersections satisfy this model:
    $$
      \begin{bmatrix}
      x^t \\
      y^t \\
      \theta ^ t
      \end{bmatrix}
      = 
      \begin{bmatrix}
      1 & 0 \\
      0 & 1 
      \end{bmatrix}
      \begin{bmatrix}
      x^{t - 1} \\
      y^{t - 1} \\
      \theta ^{t - 1} 
      \end{bmatrix}
      \eqno{(7)}
    $$
    \item \textit{Constant Velocity Model}:
    In this model, the vehicle is assumed to keep constant velocity along both the x and y axes. Thus the acceleration, yaw angle and yaw angle velocity are set to zero. The state vector can be simplified to $X = (x, y, v_x, v_y)^T$. Most of the moving vehicles in low-intensity traffic conditions satisfy this model:
    $$
      \begin{bmatrix}
      x^t \\
      y^t \\
      v_x^t \\
      v_y^t 
      \end{bmatrix}
      = 
      \begin{bmatrix}
      1 & 0 & \Delta t & 0 \\
      0 & 1 & 0 & \Delta t \\
      0 & 0 & 1 & 0 \\
      0 & 0 & 0 & 1 \\
      \end{bmatrix}
      \begin{bmatrix}
      x^{t - 1} \\
      y^{t - 1} \\
      v_x^{t - 1} \\
      v_y^{t - 1}
      \end{bmatrix} 
      \eqno{(8)}
    $$
    \item \textit{Constant Acceleration Model}: In this model, the heading angle of the vehicle may not equal the orientation of the vehicle velocity. Those vehicles that are changing lane or turning satisfy this model:
    $$
      X^t
      = 
      \begin{bmatrix}
      1 & 0 & \Delta t & 0 & \frac{1}{2}\Delta t^2 & 0  & 0 & 0\\
      0 & 1 & 0 & \Delta t & 0 & \frac{1}{2}\Delta t^2 & 0 & 0\\
      0 & 0 & 1 & 0 & \Delta t & 0 & 0 & 0\\
      0 & 0 & 0 & 1 & 0 & \Delta t & 0 & 0\\
      0 & 0 & 0 & 0 & 1 & 0 & 0 & 0\\
      0 & 0 & 0 & 0 & 0 & 1 & 0 & 0\\
      0 & 0 & 0 & 0 & 0 & 0 & 1 & 0\\ 
      0 & 0 & 0 & 0 & 0 & 0 & 0 & 1\\
      \end{bmatrix}
       X^{t-1}
      \eqno{(9)}
    $$

\end{enumerate}

KF is one of the most powerful recursive state estimators in robotics and autonomous vehicles. However, it has the limitation that it requires accurate prior knowledge of the vehicle model and noise co-variances, which are difficult to obtain due to the dynamic characteristics of the surrounding environment of the autonomous vehicle. To further improve the performance of vehicle tracking, we apply a MMA step to choose the best-fit model. The estimation residual of the measurement from each filter reflects whether a particular KF with its corresponding vehicle model is the best. In (10), we calculate the co-variance of the pre-fit residual. Based on the pre-fit residual $r_t$ and co-variance of the pre-fit residual $S_t$, we calculate the likelihood of each filter using (11) and the normalized score for each KF using (12) \cite{mmae}. The filter with the highest score will be the best selected tracker. The whole process is summarized in 

\begin{figure*}[h]
      \centering
      \includegraphics[scale=0.625]{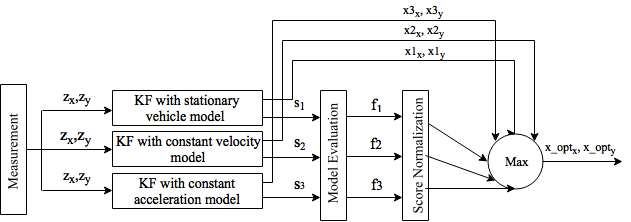}
      \caption{Best KF selection based on multiple model association}
      \label{fig:2}
\end{figure*}

$$
S_t = HPH^T 
\eqno{(10)}
$$
$$
f_n(r_t) = \frac{1}{\sqrt{(2\pi)^{\frac{m}{2}}|S_t|}}e^{-\frac{1}{2}(r_t^TS_{t}^{-1}r_t)}
\eqno{(11)}
$$
$$
p_n(t) = \frac{f_n(r_t)p_n(t - 1)}{\sum_{j = 1}^{N}f_j(r_t)p_j(t - 1)}
\eqno{(12)}
$$

\section{EXPERIMENTAL RESULTS}\label{sec:result}
CMU's autonomous vehicle platform is a retrofitted Cadillac SRX equipped with six IBEO four-layer LIDAR. The six IBEO LIDAR give a 360-degree field-of-view of the environment around the vehicle \cite{cmuplatform}. We drove the platform around the main campus of CMU in Pittsburgh, which is an urban area with congestion. We created a manually labeled dataset with vehicle heading angle, vehicle size and tracker ID. The dataset will be released for academic research along with this paper. The dataset is summarized in Table \ref{table:dataset}. 
\chiyu{
There are 5353 observations of cars from 1368 frames. Within these observations, 60 cars are fully tracked from their appearance in the field of view to their disappearance. They are manually segmented, registered and labeled. The tracked vehicles are fitted by rectangular bounding boxes with heading orientation and location of nearest corner of tracked vehicles. The point clusters of the tracked cars are not only L-shapes, there are also bumper (rear)-only or side-only observations which result in a line of LIDAR points. The proposed approach does not require the knowledge of shapes of clusters, thus the whole trajectories of tracked cars are fully used in the experiments. The dataset will be further enhanced with more vehicles and observations; the current results are tested on part of the dataset. }{}
\begin{table}[tp]
\caption{Summary of the dataset.}
\label{Table_dataset}
\begin{center}
\begin{tabular}{|c|c|}
\hline
Category & Num\\
\hline
Tracker & 60 \\
\hline
Vehicle observation & 5353 \\
\hline
frame   & 1368 \\
\hline
Observation / tracker & 89 \\
\hline
Oberservation / frame & 4 \\
\hline
L-shape view & 5037  \\
\hline 
Side view    & 115 \\
\hline
Rear view    & 201 \\
\hline
\end{tabular}
\end{center}
\label{table:dataset}
\end{table}
\subsection{Vehicle Heading Angle Estimation}
\begin{figure}
      \centering
    \begin{subfigure}[t]{0.45\linewidth}
      \centering
      \includegraphics[width=1\linewidth]{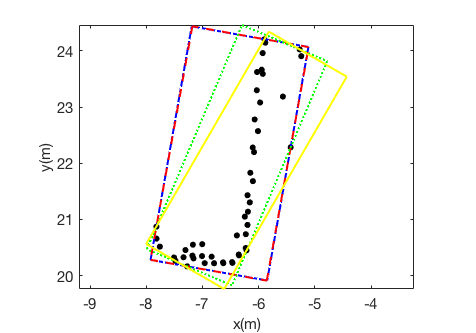}
      \caption{Case 1} \label{fig:1-case1}
    \end{subfigure}
    \hfill
    \begin{subfigure}[t]{0.45\linewidth}
        \centering
        \includegraphics[width=1\linewidth]{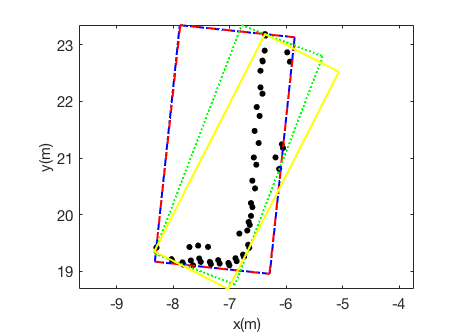} 
        \caption{Case 2} \label{fig:1-case2}
    \end{subfigure}
    \hfill
    \begin{subfigure}[t]{0.45\linewidth}
        \centering
        \includegraphics[width=1\linewidth]{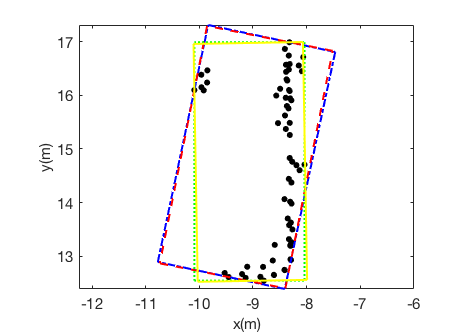} 
        \caption{Case 3} \label{fig:1-case3}
    \end{subfigure}
    \hfill
    \begin{subfigure}[t]{0.45\linewidth}
        \centering
        \includegraphics[width=1\linewidth]{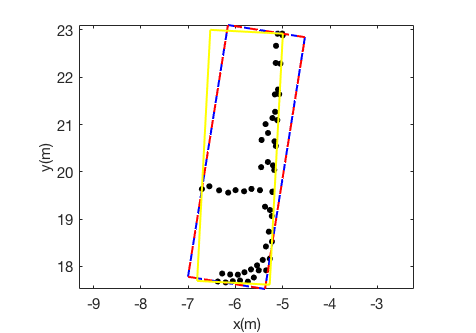} 
        \caption{Case 4} \label{fig:1-case4}
    \end{subfigure}
    \caption{Sample results of L-shape fitting using different algorithms. Yellow boxes come from the area criteria, green boxes from the closeness criteria, red boxes from the variance criteria and blue boxes from T-linkage RANSAC-based fitting.}
    \label{fig:3}
\end{figure}

The accuracy of the vehicle heading angle estimation is evaluated based on our manually labeled dataset. The samples of the fitting results from our methods are shown in Fig. \ref{fig:3}. In case 1 and case 2 in Fig. \ref{fig:1-case1} and Fig. \ref{fig:1-case2}, together with the variance criterion, our proposed T-linkage RANSAC-based method gives a better heading angle estimation than the area and closeness criteria. However, the fitting results w.r.t. these criteria were sensitive to distorting observations, such as the rear-view mirror on the side of the vehicles. In contrast, in case 3 and case 4 in Fig. \ref{fig:1-case3} and Fig. \ref{fig:1-case4}, the area criterion and closeness criterion have a better prediction than the variance criterion and T-linkage RANSAC-based method. In case 3, the two worse methods are influenced by the sub-cluster at the left-top of the cluster. In case 4, the variance and T-linkage RANSAC-based method are influenced by the back window of the vehicle, but the area and closeness criteria give a better estimation. In summary, a single criterion cannot properly qualify the estimation in all scenarios. Therefore, our best selection L-shape fitting pipeline serves as a cure to this problem. 
TABLE \ref{table_2} shows the comparison of the performance among different criteria and methods. Taking advantage of our best-fitting selection pipeline, we have a better standard deviation and mean of absolute heading error which is 1.2147, compared to 1.6162 which is the best of four criteria. Even though T-linkage is a random-sampling based method, the variance of the mean and standard deviation is less than $0.1$, which indicates the stability and robustness of the approach.

By selecting the best estimate among fitting criteria we also eliminate large orientation errors. In the single-criterion rectangle fitting methods, even the best performance variance criterion will have a large error, more than $6 \degree$ in heading angle estimation, which is not safe enough for ego-vehicle planning. TABLE \ref{table_3} shows that by using the proposed method, 99.3\% of the errors are less than $5 \degree$ and 86\% of the errors are within $2 \degree$. Our approach eliminates the large errors and improves the stability of the perception system.

\begin{table}[h]
\caption{Heading Angle Error Comparison between Different Fitting Methods}
\label{table_2}
\begin{center}
\begin{tabular}{|c|c|c|c|c|}
\hline
 & \multicolumn{2}{c|}{Real Error ($\theta - \theta_g$)} & \multicolumn{2}{c|}{Absolute Error $|\theta - \theta_g|$}\\
\hline
Method & Mean (deg) & STD (deg) & Mean (deg) & STD (deg)\\
Area   & 0.6483 & 14.7989 & 12.3310 & 8.1437 \\
Closeness & 0.0069 & 3.6467 & 2.0069 & 3.0402 \\
Variance & -0.1517 & 2.1869 & 1.4759 & 1.6162 \\
T-linkage & 0.3586 & 2.3204 & 1.5586 & 1.7515 \\
Best selection & -0.223  & -0.248 & 1.3517  & 1.2147 \\
\hline
\end{tabular}
\end{center}
\end{table}

\begin{table}[h]
\caption{Distribution of Absolute Heading Angle Error}
\label{table_3}
\begin{center}
\scalebox{0.95}{\begin{tabular}{|c|c|c|c|c|c|c|c|}
\hline
Error (deg) & = 0 & $\leq 1$ & $\leq 2$ & $\leq 3$ & $\leq 4$ & $\leq 5$\\
\hline
Variance & 22.1\% & 60.7\% & 83.4\% & 91.0\% & 95.9\% & 97.2\% \\
\hline
Best selection & 23.4\% & 60.0\% & 86.9\% & 96.5\% & 96.5\% & 99.3\% \\
\hline
\end{tabular}}
\end{center}
\end{table}

\subsection{Vehicle Tracking} 

\begin{figure}
      \centering
    \begin{subfigure}[t]{0.8\linewidth}
      \centering
      \includegraphics[width=1\linewidth]{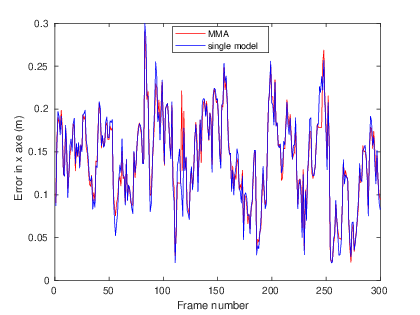}
      \caption{Absolute error of X axe} \label{fig:mmae-case1}
    \end{subfigure}
    \hfill
    \begin{subfigure}[t]{0.8\linewidth}
        \centering
        \includegraphics[width=1\linewidth]{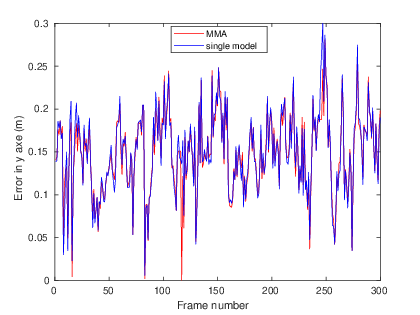} 
        \caption{Absolute error of Y axe} \label{fig:mmae-case2}
    \end{subfigure}
    \caption{Vehicle trajectory estimation result. The red line indicates the absolute trajectory prediction error of Multiple Model Association-based method. The blue line indicates the absolute trajectory prediction error of single model.}
    \label{fig:mmae}
\end{figure}

To evaluate the performance of the vehicle tracking, we first compare the accuracy of vehicle location prediction in x, y coordinates and along the vehicle motion trajectory. The pair of results comparing the vehicle position estimation result of the single-model method and position estimation applying MMA is shown in Fig. \ref{fig:mmae}. In both Fig. \ref{fig:mmae-case1} and Fig. \ref{fig:mmae-case2}, the MMA-based tracking method outperforms the single-model tracking method, giving a slightly accurate trajectory prediction of vehicles, detailed comparison is shown in TABLE \ref{table_3}.

\begin{table}[t!]
\renewcommand{\arraystretch}{1}
\caption{Performance comparison of single model trajectory estimation and MMA based trajectory estimation on stationary and moving vehicle.}
\label{tab:mmae}
\centering
\scalebox{0.9}{
\begin{tabular}[t!]{cccccc}
\toprule

Methods  &  state & x-axis(m) & x-axis(m) & y-axis(m) & y-axis(m) \\
         &                &   mean   &   STD     & mean  &    STD      \\
\cmidrule(r){1-1}\cmidrule(lr){2-2}\cmidrule(lr){3-3}\cmidrule(lr){4-4} \cmidrule(lr){5-5}\cmidrule(lr){6-6}\\

single model & stationary & 0.1263 & 0.0602 & 0.1428 & 0.0571\\
             & moving & 0.2449 & 0.1725 & 0.1918 & 0.1188\\
\midrule
 MMA         & stationary & 0.1239 & 0.0570 & 0.1376 & 0.0544\\
             & moving & 0.2384 & 0.1397 & 0.1626 & 0.1201\\
\bottomrule
\end{tabular}
}
\vspace{-0.6 cm}
\end{table}

In Table \ref{tab:mmae}, we compare the performance of the proposed method with the single-model method in detail. We calculate the average of the absolute error in x, y coordinates over different types of vehicle state. The MMA algorithm slightly improves the accuracy of vehicle position estimation with less mean and variance in error. We achieve 0.123m error for stationary vehicles in the x-axis and 0.137m in the y-axis. For moving vehicles, we achieve 0.238m error in the x-axis and 0.162m error in the y-axis. The MMA-based method will choose the best-fit model adaptively at each frame given the co-variance of the residual error. It solves the problem of inaccuracy of the vehicle model and system noise assumption, which reduce the bias of the selected model. It also gives the algorithm more ability to deal with the dynamics of surrounding vehicles. 

\section{CONCLUSIONS}\label{sec:conclusion}

In this paper, we improve previously proposed search-based L-shape fitting algorithms using T-linkage RANSAC. This novel method gets rid of various kinds of noise caused by low-cost sensors, which influence the heading angle estimation. We also improve the conventional vehicle tracking method based on Kalman filter using multiple model association. The tracking trajectories are more smooth and more dependable. Along with the paper, a manually labeled low-cost LIDAR dataset is released. Further work will be to accurately estimate the vehicle geometry in this low-density, high-occlusion and high-deformation dataset.





\section*{ACKNOWLEDGMENT}
The authors would like to thank Wenda Xu for the discussion of the LIDAR clustering and segmentation.


\bibliographystyle{IEEEtran}
\bibliography{auto}

\begin{thebibliography}{10}
\providecommand{\url}[1]{#1}
\csname url@samestyle\endcsname
\providecommand{\newblock}{\relax}
\providecommand{\bibinfo}[2]{#2}
\providecommand{\BIBentrySTDinterwordspacing}{\spaceskip=0pt\relax}
\providecommand{\BIBentryALTinterwordstretchfactor}{4}
\providecommand{\BIBentryALTinterwordspacing}{\spaceskip=\fontdimen2\font plus
\BIBentryALTinterwordstretchfactor\fontdimen3\font minus
  \fontdimen4\font\relax}
\providecommand{\BIBforeignlanguage}[2]{{%
\expandafter\ifx\csname l@#1\endcsname\relax
\typeout{** WARNING: IEEEtran.bst: No hyphenation pattern has been}%
\typeout{** loaded for the language `#1'. Using the pattern for}%
\typeout{** the default language instead.}%
\else
\language=\csname l@#1\endcsname
\fi
#2}}
\providecommand{\BIBdecl}{\relax}
\BIBdecl

\bibitem{darms2009obstacle}
M.~S. Darms, P.~E. Rybski, C.~Baker, and C.~Urmson, ``Obstacle detection and
  tracking for the urban challenge,'' \emph{IEEE Transactions on Intelligent
  Transportation Systems}, vol.~10, no.~3, pp. 475--485, 2009.

\bibitem{wei2013towards}
J.~Wei, J.~M. Snider, J.~Kim, J.~M. Dolan, R.~Rajkumar, and B.~Litkouhi,
  ``Towards a viable autonomous driving research platform,'' in \emph{IEEE
  Intelligent Vehicles Symposium}, 2013, pp. 763--770.

\bibitem{cmuplatform}
W.~Junqing, S.~Jarrod~M., K.~Junsung, D.~John~M., R.~Raj, and L.~Bakhtiar,
  ``Towards a viable autonomous driving research platform,'' in
  \emph{Intelligent Vehicles Symposium, 2013. Proceedings. IEEE}.\hskip 1em
  plus 0.5em minus 0.4em\relax IEEE, 2013, pp. 763--770.

\bibitem{Petrovskaya2009}
A.~Petrovskaya and S.~Thrun, ``Model based vehicle detection and tracking for
  autonomous urban driving,'' \emph{Autonomous Robots}, vol.~26, no.~2, pp.
  123--139, Apr 2009.

\bibitem{maclachlan2006tracking}
R.~MacLachlan and C.~Mertz, ``Tracking of moving objects from a moving vehicle
  using a scanning laser rangefinder,'' in \emph{IEEE Intelligent
  Transportation Systems Conference}, 2006, pp. 301--306.

\bibitem{shen2015efficient}
X.~Shen, S.~Pendleton, and M.~H. Ang, ``Efficient {L-shape} fitting of laser
  scanner data for vehicle pose estimation,'' in \emph{IEEE Conference on
  Robotics, Automation and Mechatronics}, 2015, pp. 173--178.

\bibitem{mertz2013moving}
C.~Mertz, L.~E. Navarro-Serment, R.~MacLachlan, P.~Rybski, A.~Steinfeld,
  A.~Suppe, C.~Urmson, N.~Vandapel, M.~Hebert, C.~Thorpe \emph{et~al.},
  ``Moving object detection with laser scanners,'' \emph{Journal of Field
  Robotics}, vol.~30, no.~1, pp. 17--43, 2013.

\bibitem{zhao2009moving}
H.~Zhao, Q.~Zhang, M.~Chiba, R.~Shibasaki, J.~Cui, and H.~Zha, ``Moving object
  classification using horizontal laser scan data,'' in \emph{IEEE
  International Conference on Robotics and Automation}, 2009, pp. 2424--2430.

\bibitem{zhang2017efficient}
X.~Zhang, W.~Xu, C.~Dong, and J.~M. Dolan, ``Efficient l-shape fitting for
  vehicle detection using laser scanners,'' in \emph{Intelligent Vehicles
  Symposium (IV), 2017 IEEE}.\hskip 1em plus 0.5em minus 0.4em\relax IEEE,
  2017, pp. 54--59.

\bibitem{mht2015}
K.~Chanho, L.~Fuxin, C.~Arridhana, and R.~James~M., ``Multiple hypothesis
  tracking revisited,'' in \emph{The IEEE International Conference on Computer
  Vision (ICCV)}.\hskip 1em plus 0.5em minus 0.4em\relax IEEE, 2015, pp.
  4696--4704.

\bibitem{jpda2015}
S.~Hamid~Rezatofighi, A.~Milan, Z.~Zhang, Q.~Shi, A.~Dick, and I.~Reid, ``Joint
  probabilistic data association revisited,'' in \emph{The IEEE International
  Conference on Computer Vision (ICCV)}.\hskip 1em plus 0.5em minus 0.4em\relax
  IEEE, December 2015, pp. 3047--3055.

\bibitem{MMM2017}
L.~Robert, ``Multitarget multisensor motion tracking of vehicles with vehicle
  based multilayer 2d laser range finders,''
  {https://mediatum.ub.tum.de/doc/1345626/1345626.pdf }, online.

\bibitem{1709.08517}
M.~Daniel, H.~Paul, Z.~William, and M.~Steve, ``Ladar-based vehicle tracking
  and trajectory estimation for urban driving,'' 2017.

\bibitem{Ego2008}
M.~Takeo, O.~Yoshihiro, and N.~Yoshiki, ``Ego-motion estimation and moving
  object tracking using multi-layer lidar,'' in \emph{Intelligent Vehicles
  Symposium, 2009. Proceedings. IEEE}.\hskip 1em plus 0.5em minus 0.4em\relax
  IEEE, 2009, pp. 151--156.

\bibitem{jlinkage}
T.~Roberto and F.~Andrea, ``Robust multiple structures estimation with
  j-linkage,'' in \emph{European Conference on Computer Vision}, 2008, pp.
  537--547.

\bibitem{Zuliani05themultiransac}
M.~Zuliani, C.~S. Kenney, and B.~S. Manjunath, ``The multiransac algorithm and
  its application to detect planar homographies,'' in \emph{In IEEE
  International Conference on Image Processing}, 2005.

\bibitem{Magri14t-linkage:a}
M.~Luca and F.~Andrea, ``T-linkage: A continuous relaxation of j-linkage for
  multi-model fitting,'' in \emph{In Proceedings of the IEEE Conference on
  Computer Vision and Pattern Recognition}, 2014, pp. 3954--3961.

\bibitem{Tdist}
T.~T., ``An elementary mathematical theory of classification and prediction,''
  in \emph{Internal IBM Technical Report}, 1957.

\bibitem{Reid79analgorithm}
D.~B. Reid, ``An algorithm for tracking multiple targets,'' \emph{IEEE
  Transactions on Automatic Control}, vol.~24, pp. 843--854, 1979.

\bibitem{mmae}
H.~Peter~D. and M.~Peter~s., ``Multiple-model adaptive estimation using a
  residual correlation kalman filter bank,'' \emph{IEEE Transactions on
  Aerospace and Electronic Systems}, vol.~36, no.~2, pp. 393--406, 2000.

\end{thebibliography}

\end{document}